\documentclass[letterpaper, 10 pt, conference]{ieeeconf}  

\IEEEoverridecommandlockouts                              

\usepackage{subcaption}
\usepackage{amsmath,amsfonts}
\usepackage{algorithm}
\usepackage[noend]{algpseudocode}
\usepackage{amssymb}
\usepackage{amsmath}
\usepackage{mathtools}
\usepackage{array}
\usepackage{textcomp}
\usepackage{stfloats}
\usepackage{url}
\usepackage{verbatim}
\usepackage{graphicx}
\usepackage{cite}
\usepackage{hyperref}
\usepackage[switch]{lineno}
\usepackage{xcolor}

\def\*#1{\mathbf{#1}}  
\newcommand{\bx}{\mathbf{x}}
\newcommand{\bu}{\mathbf{u}}
\newcommand{\bv}{\mathbf{v}}
\newcommand{\bw}{\boldsymbol{\omega}}
\newcommand{\bs}{\mathbf{s}}
\newcommand{\bp}{\mathbf{p}}
\newcommand{\bP}{\mathbf{P}}
\newcommand{\bM}{\mathbf{M}}
\newcommand{\bt}{\mathbf{t}}

\newcommand{\br}{\mathbf{r}}

\newcommand{\bA}{\mathbf{A}}
\newcommand{\bB}{\mathbf{B}}
\newcommand{\ba}{\mathbf{a}}
\newcommand{\bb}{\mathbf{b}}
\newcommand{\bI}{\mathbf{I}}
\newcommand{\bR}{\mathbf{R}}

\newcommand{\IGNORE}[1]{}

\newcommand{\RED}[1]{\textcolor{red}{#1}}
\newcommand{\BLUE}[1]{\textcolor{blue}{#1}}

\begin{document}%

\title{\LARGE \bf
SMF-VO: Direct Ego-Motion Estimation via Sparse Motion Fields
}

\author{\IGNORE{\RED{(arxiv ver.)~~~}}Sangheon Yang$^{1}$, Yeongin Yoon$^{2}$, Hong Mo Jung$^{2}$, and Jongwoo Lim$^{*2}$
\thanks{$^{*}$Corresponding Author}%
\thanks{$^{1}$Department of Artificial Intelligence, Hanyang University, Seoul, Korea. {
\tt \begin{scriptsize} philippine96@hanyang.ac.kr
\end{scriptsize}}
}%
\thanks{$^{2}$Department of Mechanical Engineering, Seoul National University, Seoul, Korea
{
\tt \begin{scriptsize}\{yoonyi2017;jjhm2910;jongwoo.lim\}@snu.ac.kr
\end{scriptsize}}
}%
\thanks{All authors are with the Robot Vision Lab, Seoul National University.}
}

\maketitle
\thispagestyle{empty}
\pagestyle{empty}


\begin{abstract}
Traditional Visual Odometry (VO) and Visual-Inertial Odometry (VIO) methods rely on a ``pose-centric" paradigm, which computes absolute camera poses from the local map thus requires large-scale landmark maintenance and continuous map optimization. This approach is computationally expensive, limiting their real-time performance on resource-constrained devices. 

To overcome these limitations, we introduce Sparse Motion-Field Visual Odometry (SMF-VO), a lightweight, ``motion-centric" framework. Our approach directly estimates instantaneous linear and angular velocity from sparse optical flow, bypassing the need for explicit pose estimation or expensive landmark tracking. \IGNORE{\RED}{We also employed a generalized 3D ray-based motion field formulation that works accurately with various camera models, including wide-field-of-view lenses.}

SMF-VO demonstrates superior efficiency and competitive accuracy on benchmark datasets, achieving over 100 FPS on a Raspberry Pi 5 using only a CPU. Our work establishes a scalable and efficient alternative to conventional methods, making it highly suitable for mobile robotics and wearable devices.

\end{abstract}


\section{INTRODUCTION}

Visual Odometry (VO) is a fundamental technique for estimating camera motion from visual input and is widely used in robotics, augmented/virtual reality (AR/VR), and autonomous navigation. By analyzing sequential images captured from a monocular, stereo, or RGB-D camera, it estimates the ego-motion of the camera. 
Despite substantial progress in accuracy and robustness, conventional VO systems still have several limitations, including heavy computational overhead and sensitivity to environmental changes such as lighting variations and dynamic objects. 

The main reason for the computational overhead is that traditional feature-based VO algorithms follow the ``pose-centric" paradigm, which requires explicit camera pose estimation (P3P or PnP) based on the local map. 
Since pose accuracy relies critically on map quality, this process requires tracking a large number of accumulated 3D landmarks over multiple keyframes and performing local bundle adjustment to keep the local map accurate, leading to high computational costs and makes real-time operation challenging on resource-constrained platforms.


Visual-inertial odometry (VIO) incorporates inertial measurements to improve reliability in visually degraded conditions. However, this integration introduces new challenges, including multi-sensor calibration, pose drift due to IMU bias, and increased algorithmic complexity. Additionally, pose-centric VIO frameworks inherit the computational burdens of VO on landmark tracking and continuous optimization.

\begin{figure}[tb!]
    \centering
    \begin{subfigure}[b]{0.40\textwidth}
        \centering
        \includegraphics[width=\linewidth]{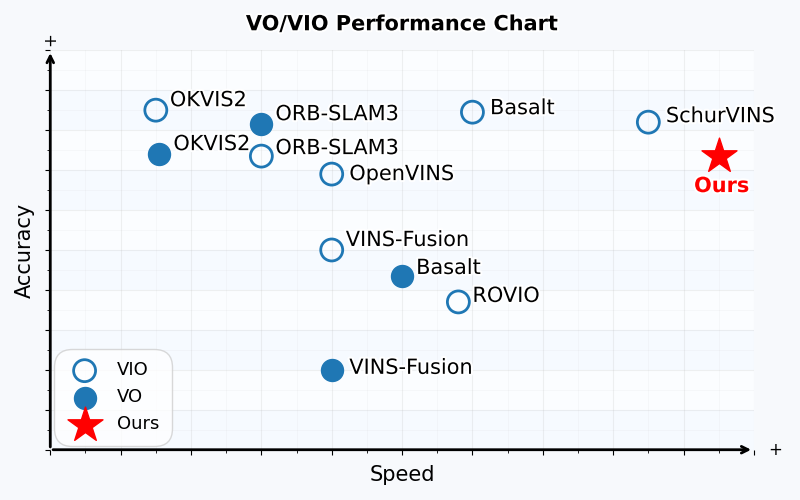}
    \end{subfigure}
    \hfill
    \begin{subfigure}[b]{0.45\textwidth}
        \begin{center}
        \resizebox{\linewidth}{!}{%
        \begin{tabular}{l l c c c}
        \hline
        \textbf{Method (VO/VIO)} & Key component of state estimation & Execution speed & Accuracy \\
        \hline
        {(IMU dead reckon)} & (Acceleration \& Angular velocity) & ($\star\star\star\star\star$) & ($\star$) \\
        \textbf{\textit{{Ours}}} & \textbf{Linear \& Angular velocity} & $\star\star\star~\star$ & $\star\star\star$ \\
        {Filter-based} & Position \& Rotation (+ velocity) & $\star\star\star$ & $\star~\star$ \\
        {Optimization-based} & Position \& Rotation & $\star~\star$ & $\star\star\star~\star$ \\
        \hline
        \end{tabular}
        }
        \end{center}
    \end{subfigure}
    \caption{An overall comparison of the accuracy and speed of VO and VIO algorithms. Our algorithm directly estimates camera motion from the visual motion field. Thus, it runs much faster than conventional VO algorithms without sacrificing accuracy. }
    \label{fig:combined_figure}
\end{figure}

To address these challenges, we introduce Sparse Motion-Field Visual Odometry (SMF-VO), a lightweight, high-speed framework that directly estimates ego-motion from motion field equations. Our novel ``motion-centric" approach overcomes the limitations of traditional VO and VIO methods by directly calculating the camera's instantaneous linear and angular velocity from sparse optical flow. This method eliminates the need for explicit pose estimation, such as P3P, and the computationally expensive multi-frame optimization required for large-scale landmark maintenance. SMF-VO instead solves a compact per-frame optimization problem using only sparse keypoints in each frame, making it highly efficient for real-time deployment on low-power embedded devices.

{Building upon the classical motion field, we employ a generalized 3D ray-based formulation by representing image points as normalized rays. This approach removes the limitations of conventional 2D pixel-based, pinhole-model equations, allowing the resulting SMF-VO framework to accommodate any camera model. This is crucial for handling cameras with varied fields of view and nonlinear projection, like fisheye lenses in robotics.}

Through extensive experimentation with various benchmark datasets, including EuRoC, KITTI, and TUM-VI, we demonstrate that SMF-VO is significantly more computationally efficient than conventional VO methods while maintaining comparable accuracy. The system achieves over 100 frames per second (FPS) on a Raspberry Pi 5 using only a quad-core 2.4 GHz CPU. By pioneering a motion-centric ego-motion estimation paradigm, SMF-VO provides a scalable and efficient alternative to conventional VO, making it a strong candidate for practical, real-life applications in mobile robotics, AR/VR, and wearable devices.


\IGNORE{Our work introduces SMF-VO, a novel visual odometry framework that makes three main contributions.}
\IGNORE{\RED{edited}}
{Our work introduces SMF-VO, a novel visual odometry framework that offers three key features.}
\begin{itemize}
    \item \textbf{A New Motion-Centric Paradigm}: We developed a new visual odometry approach that directly estimates camera ego-motion as linear and angular velocity, instead of using traditional position-based methods. This allows for high-efficiency real-time performance without sacrificing accuracy.
    \IGNORE{\item \RED{\textbf{Generalizable Camera Support}: We developed a generalized 3D ray-based motion field formulation that works accurately with various camera models, especially wide field-of-view systems like fisheye lenses. This overcomes the limitation of traditional 2D coordinate-based methods, which are only compatible with the pinhole camera model.}}
    \item \IGNORE{\RED{edited}}{\textbf{Generalizable Camera Support}: We achieve camera model generalizability by employing a generalized 3D ray-based motion field formulation. This existing concept, when applied to our framework, works accurately with various models, including wide field-of-view systems like fisheye lenses, overcoming the limitations of traditional pinhole-only 2D methods.}
    \item \textbf{Superior Real-World Performance}: We validated our system on standard datasets such as EuRoC, KITTI, and TUM-VI. Our algorithm achieved over 100 FPS on a Raspberry Pi 5 using only a CPU, demonstrating its superior efficiency and robustness. This proves its suitability for real-time applications on resource-constrained devices.
\end{itemize}



\section{RELATED WORKS}

\noindent\textbf{Visual and Visual Inertial Odometry.}
Traditional Visual Odometry (VO) methods can be categorized into feature-based and direct approaches. 
Feature-based methods \cite{Davison03, Mouragnon06, Klein07, Mur-Artal15} track or match sparse keypoints across frames and use geometric constraints for motion estimation. These methods are effective in well-textured environments, but suffer from unreliable feature correspondences in low-texture or dynamic scenes. 
Stereo VO \cite{Geiger11, Mur-Artal17} incorporates depth information to compute metric motion without scale drift, but remains sensitive to illumination changes and occlusions.
Direct methods \cite{Engel14, Forster16, Engel17} estimate motion by minimizing photometric error over pixel intensities rather than relying on feature correspondences. While effective in textureless environments, they are computationally expensive and highly sensitive to lighting variations and motion blur.
Optimization-based approaches, such as Bundle Adjustment (BA) and Factor Graph-based SLAM \cite{Strasdat11, Kummerle11}, further refine motion estimates but impose a high computational burden, limiting their real-time feasibility on embedded systems.
VIO methods integrate an IMU with a camera to enhance robustness in visually degraded environments.
Filtering-based methods such as MSCKF \cite{Mourikis07}, ROVIO \cite{Bloesch17}, and SchurVINS \cite{Fan_2024_CVPR} use recursive filtering to propagate state estimates, while optimization-based methods such as VINS-Mono \cite{Qin18}, Basalt \cite{Usenko19}, and OKVIS2 \cite{Leutenegger22} employ non-linear optimization for improved accuracy.
Despite their advantages, VIO approaches introduce additional complexities, including sensor drift, calibration sensitivity, and increased computational cost.

Unlike these methods, SMF-VO directly estimates velocity using motion field equations from sparse optical flow, eliminating the need for explicit pose estimation via P3P algorithm or inertial sensing. By formulating the problem as a compact per-frame optimization, it reduces computational overhead enabling real-time operation on low-power embedded systems.

\noindent\textbf{Ego-motion estimation with Motion Field.}
Motion field constraints have been widely studied in 3D computer vision \cite{LonguetHiggins80, Trucco98}, describing the relationship between optical flow and camera motion in 3D.
Based on this relationship, the motion field has been used to compute ground truth optical flow for LiDAR points in event camera datasets \cite{Zhu18, Zhu18RSS, Gehrig21, Gehrig21ERAFT}. 
Recent studies have also explored their use in event-based vision, using high-speed asynchronous data to estimate motion.
Self-supervised learning approaches \cite{Zhu19, Ye20} first learn to estimate ego-motion and depth from event camera data and then utilize the motion field equation to compute optical flow.
Shiba et al. \cite{Shiba24} parameterized optical flow using motion field equations for contrast maximization in event cameras, while Lu et al. \cite{Lu24} fused motion field constraints with IMU data to recover velocity.

These methods, which either rely on event-based sensors or require IMU fusion, use the motion field equations to (1) apply them in parameterization processes for optimization, (2) estimate linear velocity only, or (3) estimate depth and ego motion jointly. 
In contrast, SMF-VO applies motion field constraints to standard RGB cameras to achieve direct ego-motion estimation without the need for additional sensors or computationally expensive optimization.

\noindent\textbf{Computational Efficiency in Visual Odometry.}
Efficient real-time VO is crucial for deployment in resource-constrained systems.  
Optimization-based SLAM frameworks such as ORB-SLAM3 \cite{Campos21} and Basalt \cite{Usenko19} achieve high accuracy but require multi-threaded execution and substantial computational power.
Lightweight VO systems such as SVO \cite{Forster16} and DSO \cite{Engel17} improve efficiency by reducing feature tracking overhead, but they still rely on photometric constraints, making them susceptible to lighting variations.
SMF-VO achieves over 100 FPS speed without any GPU or hardware acceleration. This makes it a practical alternative for real-time applications on embedded platforms.


\section{Methodology}
\begin{figure}[tb!]
    \centering
     \includegraphics[width=\linewidth]{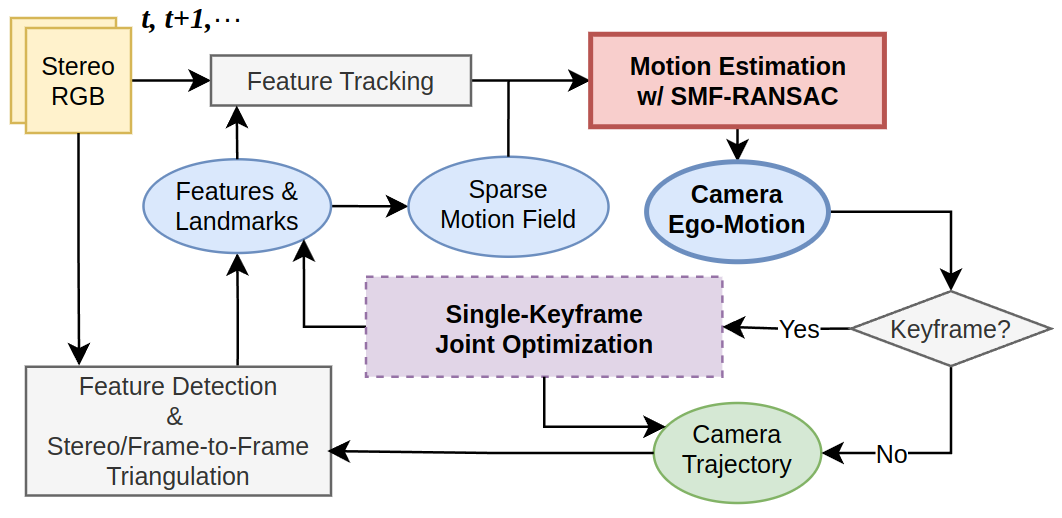}
    \caption{Overview of the framework we propose.}
    \label{fig:sys_overview}
\end{figure}

The 3D motion of the camera can be computed from the motions of the projected points at known 3D positions~\cite{LonguetHiggins80,Trucco98}. We briefly summarize the motion-field formulation below. 
The bold symbols represent vectors, and the subscript represents the corresponding coordinate in them, e.g., for $\bx = [a~b~c]^\top$, $\bx_x = a$, $\bx_y = b$, and $\bx_{(x,y)} = [a~b]^\top$. The dot above a vector represents the temporal derivative of the vector, i.e. $\dot\bx = \frac{d}{dt}\bx$.

\subsection{3D Camera Motion from Sparse Motion Field} \label{subsec:2DMF}

For a 3D point in the camera coordinate frame, $\bP = [X~Y~Z]^\top$, the corresponding point projected onto the image plane at focal length $f$ can be expressed as 
\begin{equation}\label{eq_proj}
\bp = f\;({\bP}/{Z}) \;\;:= [x~ y~ f]^\top.\\
\end{equation}
When the camera moves with the linear velocity $\bv$ and angular velocity $\bw$, the movement (velocity) of the point $\bP$ is described by
\begin{equation} \label{eq_motion_3D} 
\dot\bP = -\bv - \bw \times \bP.
\end{equation}
Differentiating both sides of \eqref{eq_proj} with respect to time gives 
\begin{equation}\label{eq_motion_2D}
\IGNORE{\dot\bp = f\frac{Z\dot\bP - \dot Z\bP}{Z^2},\\}
\dot\bp = f(Z\dot\bP - \dot Z\bP)/Z^2,\\
\end{equation}
establishing the relationship between the velocity of the 3D point $\bP$ and the velocity of $\bp$ projected on the image plane. 
Substituting $\dot\bP$ and $\dot Z$ in \eqref{eq_motion_2D} with \eqref{eq_motion_3D}, we obtain the following. 
\begin{equation}\label{eq_motion_field}
\begin{aligned}
\IGNORE{
    \dot\bp_x &= \frac{{\bv_z}x - {\bv_x}f}{Z} - {\bw_y}f + {\bw_z}y + \frac{{\bw_x}xy}{f} - \frac{{\bw_y}x^2}{f} \\
    \dot\bp_y &= \frac{{\bv_z}y - {\bv_y}f}{Z} - {\bw_x}f + {\bw_z}x + \frac{{\bw_y}xy}{f} -  \frac{{\bw_x}y^2}{f}
}
    \dot\bp_x &= {({\bv_z}x - {\bv_x}f)/Z} - {\bw_y}f + {\bw_z}y + ({{\bw_x}xy} - {{\bw_y}x^2})/f \\
    \dot\bp_y &= {({\bv_z}y - {\bv_y}f)/Z} - {\bw_x}f + {\bw_z}x + ({{\bw_y}xy} -  {{\bw_x}y^2})/f
\end{aligned}  
\end{equation}

Let $\bu=\dot\bp_{(x,y)}$ be the pixel velocity at $\bp_{(x,y)}$. 
From \eqref{eq_motion_field}, we can obtain two constraints on the 3D camera motion $\bs := [\bw^\top ~ \bv^\top]^\top \in \mathbb{R}^6$ from each pixel movement observed in the image, \vspace{-.5em}
\begin{equation}\label{eq_linear_mf}
\IGNORE{\bu = M(\bp,Z)\;\bs, \;\;\textrm{where}}
\bu = \begin{bmatrix}\bA(\bp)&\bB(\bp)/Z\end{bmatrix}\;\bs, \;\;\textrm{where}
\end{equation}
{\small \[
\setlength\arraycolsep{3pt}
\bA(\bp) = 
\begin{bmatrix} {xy}/{f} & -f-{x^2}/{f} & y 
\\ f + {y^2}/{f} & -{xy}/{f} & -x \end{bmatrix}, 
\bB(\bp) = 
\begin{bmatrix} -f & 0 & x \\ 0 & -f & y \end{bmatrix}
\IGNORE{ M{(\bp,Z)} = \begin{bmatrix} 
-{f}/{Z} & 0 & {x}/{Z} & {xy}/{f} & -f-{x^2}/{f} & y \\
0 & -{f}/{Z} & {y}/{Z} & f + {y^2}/{f} & -{xy}/{f} & -x \\
\end{bmatrix} }
\]}

From $n$ observed pixel velocities at the current frame, the 3D camera velocity can be estimated by solving the (over-constrained) linear system 
\begin{equation}\label{eq_linear_mf2}
\bar\bu \;=\; \mathbf{W}\;\bs, \;\;\textrm{where}
\end{equation}
\[
\setlength\arraycolsep{2pt}
\bar\bu = \begin{bmatrix}
\bu^{1} \\
\bu^{2} \\
\vdots\\
\bu^{n} \\
\end{bmatrix} \in \mathbb{R}^{2n},\;
\mathbf{W} = \begin{bmatrix}
\bA(\bp^{1}) & \bB(\bp^{1})/Z^{1}\\
\bA(\bp^{2}) & \bB(\bp^{2})/Z^{2}\\
\vdots & \vdots\\
\bA(\bp^{n}) & \bB(\bp^{n})/Z^{n}\\
\end{bmatrix} \in \mathbb{R}^{2n \times 6}.
\]
The depth $Z$ of each point in \eqref{eq_linear_mf2} can be obtained from the location of its corresponding landmark, which is estimated via stereo or frame-to-frame triangulation during motion estimation. 
For observations $n \geq 3$, we can solve the linear system above as
\begin{equation}\label{eq_solve_linear_mf}
\hat\bs = {[ \hat{\bw}^\top~~ \hat{\bv}^\top]}^\top = \left( \mathbf{W}^\top \mathbf{W} \right)^{-1} \mathbf{W}^\top \bar\bu \;.
\end{equation}

\subsection{Motion Field with Normalized Ray}\label{subsec:3dray}

\IGNORE{The motion field equation based on 2D pixel velocities, as described in Section~\ref{subsec:2DMF}, is limited by its assumption of an ideal pinhole camera model without distortion. However, in practical applications, cameras often employ wider Field-of-View (FoV) models, such as fisheye lenses or distorted pinhole models with calibration parameters. To address this limitation, we extend the original motion field equation to operate on 3D rays emanating from the camera's optical center, rather than relying on pixel locations in the image plane.}
\IGNORE{\RED{edited}}{The 2D pixel-based motion field (Section \ref{subsec:2DMF}) is restricted to the ideal pinhole camera model, which prohibits its use with practical wide Field-of-View (FoV) systems like fisheye lenses. To address this, we derive a generalized 3D ray-based motion field equation operating on rays from the optical center, removing reliance on 2D pixel coordinates.}

For a 3D point $\bP$ in camera coordinate, the corresponding ray $\br$ can be expressed as
\begin{equation}\label{eq_proj_3dray}
\br = {\bP}/d \;:= [x~ y~ z]^\top,~~ \text{where}~~d = ||\bP||.\\
\end{equation}
Differentiating both sides of \eqref{eq_proj_3dray} with the same way of \eqref{eq_proj} to \eqref{eq_motion_2D} gives
\begin{equation}\label{eq_motion_2D_3dray}
\begin{aligned}
\dot\br &= {(d\dot\bP - \dot d \bP)}/{d^2},  \text{where} ~~ \dot{d} = {(\bP\cdot\dot{\bP})}/{\|\bP\|} = \br\cdot\dot{\bP}, \\
\dot{\br} &= \frac{d\,\dot{\bP}-(\br\cdot\dot{\bP})\,\bP}{d^2} \;=\; \frac{d\,\dot{\bP}-(\bP\br^\top)\dot{\bP}}{d^2} \;=\; \frac{\bI - \br\br^\top}{d}\dot{\bP}, 
\end{aligned}
\end{equation}
establishing the relationship between the velocity of the 3D
point $\bP$ and the flow (velocity) of corresponding ray $\br$, that can be computed as $\dot{\br} = \br^{cur} - \br^{prev}$ .

Substituting $\dot\bP$ in \eqref{eq_motion_2D_3dray} with \eqref{eq_motion_3D}, we obtain the following. 
\IGNORE{
\begin{equation}\label{eq_motion_field_extension_3dray}
\begin{aligned}
\dot{\br} &= (1/d)~(\br\br^\top - \bI) (\bv + \bw\times\bP) \\
&= (1/d)~ (\br\br^\top - \bI) (-[\bP]_\times\bw + \bv) \\
&= (1/d)~((\bI - \br\br^\top)[\bP]_\times\bw +(\br\br^\top - \bI)\bv)\\
&= (\bI - \br\br^\top)[\br]_\times\bw +(1/d)(\br\br^\top - \bI)\bv\\
&= [\br]_\times\bw +(1/d)(\br\br^\top - \bI)\bv
\end{aligned}  
\end{equation}
}
\begin{equation}\label{eq_motion_field_extension_3dray}
\begin{aligned}
\dot{\br} &= (\bI - \br\br^\top) (-\bv - \bw\times\bP) / d \\
&= (\bI - \br\br^\top)[\bP]_\times\bw / d +(\br\br^\top - \bI)\bv / d\\
&= (\bI - \br\br^\top)[\br]_\times\bw + (\br\br^\top - \bI)\bv / d\\
&= [\br]_\times\bw + (\br\br^\top - \bI)\bv/d
\end{aligned}  
\end{equation}
\IGNORE{\BLUE{edited}}{This derived equation is consistent with existing formulations in the literature, specifically matching the motion field equation for spherical projection presented in \cite{hui2013determining} under the condition $f=1$. This validates that our general formulation yields an equivalent result for the special case of projection onto the unit sphere.}

From \eqref{eq_motion_field_extension_3dray}, each ray movement provides three constraints on the 3D camera motion form each ray movement, $\dot\br_x, \dot\br_y, \dot\br_z$; however, these constraints are rank-deficient with an effective rank of two. Given $n$ observed ray velocities at the current frame, the 3D camera velocity can be estimated by solving the resulting (over-constrained) linear system in the same manner as described in Section~\ref{subsec:2DMF}.
\begin{equation}\label{eq_linear_mf_3dray}
\begin{aligned}
\grave\bu \;&=\; \mathbf{M}\;\bs, \;\;\textrm{where} \\
\grave\bu &= [\dot\br^{1\top}~~\dot\br^{2\top}~~\dots~~\dot\br^{n\top} ]^\top \in \mathbb{R}^{3n},\;\\
\mathbf{M} &= \begin{bmatrix}
[\br^{1}]_\times & (\br^{1}\br^{1\top} - \bI) / d^{1}\\
[\br^{2}]_\times & (\br^{2}\br^{2\top} - \bI) / d^{2}\\
\vdots & \vdots\\
[\br^{n}]_\times & (\br^{n}\br^{n\top} - \bI) / d^{n}\\
\end{bmatrix} \in \mathbb{R}^{3n \times 6}
\end{aligned}
\end{equation}

The Euclidean depth $d$ of each point in \eqref{eq_linear_mf_3dray} can be obtained by stereo depth estimation \cite{Hartley03}. For observations $n \geq 3$, we can solve the linear system above as
\begin{equation}\label{eq_solve_linear_mf_3dray}
\hat\bs = {[ \hat{\bw}^\top~~ \hat{\bv}^\top]}^\top = \left( \mathbf{M}^\top \mathbf{M} \right)^{-1} \mathbf{M}^\top \grave\bu \;.
\end{equation}

Compared to the widely used P3P(PnP) algorithm \cite{Gao03} in visual odometry, our motion estimation method solves a simple $6 \times 6$ linear least squares problem of the form $\bA\bx = \bb$ to compute the instantaneous camera motion ($\bw$, $\bv$). In contrast, the P3P algorithm requires solving a quartic polynomial to estimate the pose $[\bR | \bt]$, which is computationally heavier due to the need to disambiguate multiple solutions and perform backsubstitution. 

\subsection{Robust Estimation using RANSAC}

In the previous section, we formulate the linear solver for 3D camera motion from pixel ray motions. 
However, the estimated pixel (ray) motions can be noisy or inconsistent due to outliers, e.g., independently moving objects. 
To handle such cases, the RANSAC algorithm \cite{Fischler81} is commonly used, and we also use it for robust motion estimation, as in Alg ~\ref{alg:ransac}. 

\begin{algorithm}[bth]
\caption{Motion Field RANSAC Process}\label{alg:ransac}
\begin{algorithmic}[1]
\State{$\blacktriangleright\; \{\,\br^{i},\dot\br^{i},d^{i}, \bP_i \,\}_{i=1:n}$ : rays, flows, depths, landmarks } 
\IGNORE{\State{$\blacktriangleright\; {\Pi}(\bP)$ : projection returning $f(X/Z,Y/Z)^\top$}} 
\State{$\blacktriangleright\; Q, n_s, N_{max}, \gamma_0, \tau_u, \tau_\theta$ : predefined parameters}
\State{$\blacktriangleright\; \Theta(\ba,\bb)$: angle between two vectors $\ba,\bb \in \mathbb{R}^3 $}
\Function{Estimate}{ $\{j\}$ } 
    \State build $\bM$ and $\grave\bu$ for the points in $\{j\}$.
        \Comment{eq \eqref{eq_linear_mf_3dray}}
    \State solve $\hat\bs$ for $\grave\bu = \bM\hat\bs$. ($\hat\bs = [\hat\bw^\top, \hat\bv^\top]^\top$) 
        \Comment{eq \eqref{eq_solve_linear_mf_3dray}}
    \IGNORE{\State $\hat{\bR} \leftarrow \text{Exp}(\hat\bw)$
        \Comment{note $\hat\bs = (\hat\bw^\top, \hat\bv^\top)^\top$}}
    \For{all $i$ in $1:n$} find the inliers $\{i^*\}$ s.t.
        \State $\Theta\left((\br^{i} + \dot{\br}^{i}) ~,~ \text{Exp}(\hat{\bw})^{\top}
        (\bP_{i} - \hat{\bv})\right) < \tau_{\theta}$  and
        \State $||\dot{\br}^{i} - ([\br^{i}]_\times\hat\bw + (\br^{i}\br^{i\top} - \bI)\hat\bv/d^{i}) || < \tau_{u} $
        \IGNORE{\State $(\|\;\bu^{(i)} - M(\bp^{(i)},Z^{(i)})\, \hat\bs\;\| < {\tau}_u )$ and}
        \IGNORE{\State $(\|\,(\bp^{(i)}_{(x,y)}\!+\!\bu^{(i)}) - \Pi\left( \hat{R}\frac{Z^{(i)}}{f}\bp^{(i)} + \hat\bv \right) \| \!<\! \tau_\Pi )$ 
        }
    \EndFor
    \State \textbf{return} $(\,\hat\bs,\;\{i*\}\,)$
\EndFunction
\Function{Ransac}{ } 
    \If{$n < n_s$} {\;\textbf{return} $(\,\*0^6,\,\{\}\,)$} 
    \EndIf
    \State $N_{iter} \leftarrow N_{max},\; \bs^* \leftarrow \*0^6,\; I^*\leftarrow\{\} $
    \While{\#iteration $ < N_{iter}$}
        \State ${\{j\}} \leftarrow $ $n_s$ randomly selected indices in $(1:n)$
        \State {$(\,\bs,\; I\,) \;\leftarrow\;$} \Call{Estimate}{$\{j\}$}
        \If{$\;\| I \| > \| I^* \|\;$} 
            \State $\bs^* \leftarrow \bs,\; I^* \leftarrow I$
            \If{$\;\| I^* \| / n > \gamma_0$ }{ break \textbf{while} loop}
            \EndIf
            \State $ N_{iter} \leftarrow \text{min} \left(\left\lceil \frac{\log(1 - Q)}{\log(1 - \|I^*\|/n)}\right\rceil, N_{iter}\right)$
        \EndIf
    \EndWhile
    \If{$\;\|I^*\| > n_s\;$}
        \State {$(\,\bs,\; I\,) \;\leftarrow\;$} \Call{Estimate}{$\{I^*\}$}
        \If{$\;\| I \| > \| I^* \|\;$} {$\;\bs^* \leftarrow \bs,\; I^* \leftarrow I$}
        \EndIf
    \EndIf
    \State \textbf{return} $(\,\bs^*,\;I^*\,)$

\EndFunction
\end{algorithmic}
\end{algorithm}

In our framework, a set of ray velocities $\grave\bu$ is estimated using Kanade-Lucas-Tomasi Feature Tracking \cite{Baker04} between consecutive frames.

The ESTIMATE() routine computes the linear least squares solution for a subset of ray motions and depths, while the RANSAC() routine iteratively refines 3D motion estimation by selecting the best model with the most inliers.  
The key RANSAC parameters are: success probability $Q = 99.99\%$, sample size $n_s = 3$, early termination threshold $\gamma_0 = 0.9$, and inlier thresholds $\tau_u$ and $\tau_\Pi$, that $\tau_u = 2\text{sin}(\tau_\Pi/2)$ for the ray-based, and $\tau_u = f\text{tan}(\tau_\Pi)$ for the pixel-based motion field equation.  
To determine inliers from the instantaneous pixel ray motions, we evaluate the reprojection error and the motion residuals together.  

Once the loop terminates, the final 3D camera motion and inliers are re-computed using all inliers, as the best motion from the RANSAC is initially derived from only $n_s$ minimal data points.  
Contrary to the common belief that linearized formulations are suboptimal, we find that these solutions provide highly accurate camera trajectories even when simply accumulated over time.  
To further reduce the drift caused by real-world noise, as described in the next section, a simple non-linear optimization on the camera motion $(\hat\bw,\hat\bv)$ and landmark positions $\{P_i\}$ is applied.  

\subsection{Motion and Landmark Optimization} \label{sec:n-opt}

Although the 3D motion estimated by our SMF-RANSAC is generally accurate, small errors can accumulate over time, causing pose drift. Additionally, our algorithm uses landmark distances and depths from short-baseline stereo, which can be noisy, particularly for distant features. To address these issues, we incorporate a lightweight, bundle-adjustment-style, nonlinear optimization step. This process is optional, and even when used, it is highly efficient because it optimizes only the pose of the current keyframe and its associated landmarks while fixing the other keyframes. This approach is significantly more lightweight than conventional local bundle adjustment, which optimizes a much larger set of keyframes and landmarks. 

Keyframes are selected following standard practices commonly used in graph-based VO/SLAM systems. Specifically, a frame is promoted to a keyframe when the number of tracked inlier features falls below a threshold ($< \tau_{n}$, where $\tau_{n}$ denotes the desired minimum number of observed features), or based on the elapsed time and the magnitude of relative motion between the current frame and the last keyframe, quantified by $||\bR||$ and $||\bt||$.

If a frame is selected as a keyframe, a lightweight nonlinear optimization is performed to refine the motion estimate and the landmark positions as
\begin{equation}\label{ba}
\begin{aligned}
&\{[\bR|\bt]^k\},\{\bP_{l}\} = \\
&\operatorname*{argmin}_{\{[\bR|\bt]^k\},\{\bP_{l}\}} \sum_{k,l} \rho\left(\left\|\br_l^k- \frac{{\bR^k}^\top(\bP_l-\bt^k)}{\|{\bR^k}^\top(\bP_l-\bt^k)\|} \right\|\right).
\end{aligned}  
\end{equation}
All inlier landmarks ${\bP_l}$ observed in the newly selected keyframe, along with the poses of the keyframes ${[\bR|\bt]^k}$ that observe these landmarks, are jointly optimized by minimizing the reprojection ray errors between the observed and predicted rays. In our approach, only the pose of the newly selected keyframe is treated as active, while all other keyframe poses remain fixed. This design keeps the problem size small, enabling efficient optimization and maintaining high-speed execution. The robust Cauchy loss function is applied, defined as $\rho(s) = c^2 \cdot \log\left(1 + s/{c^2}\right)$, where $c$ is a tunable scale parameter.

This procedure resembles typical bundle adjustment but is restricted to only the landmarks observed in the current keyframe. Consequently, it is significantly more lightweight than conventional local bundle adjustment, enabling high-speed execution even on embedded platforms such as the Raspberry Pi.


\section{Experimental Evaluation} 

\begin{table*}[tb!]
\centering
\resizebox{\linewidth}{!}{
\begin{tabular}{| r c | r | r r r r r | r r r | r r |}
\hline
\textbf{RMSE ATE (m)} & \textbf{Method} & \textbf{Avg} & \textbf{MH01} & \textbf{MH02} & \textbf{MH03} & \textbf{MH04} & \textbf{MH05} & \textbf{V101} & \textbf{V102} & \textbf{V103} & \textbf{V201} & \textbf{V202}  \\
\hline
\textbf{SMF-VO (ours)} & stereo VO & 0.128 & 0.099 & \underline{0.039} & 0.169 & 0.208 & 0.348 & 0.063 & 0.100 & \textbf{0.088} & 0.057 & 0.114  \\
\textbf{BASALT}~\cite{VonStumberg18} & stereo VO & 0.333 & 0.072 & 0.082 & 0.116 & \underline{0.176} & \underline{0.152} & \underline{0.058} & 0.124 & 1.974 & 0.068 & 0.511 \\
\textbf{\textbf{*}ORB-SLAM3}~\cite{Campos21}  & stereo VO & \textbf{0.088} & \underline{0.048} & \textbf{0.017} & \textbf{0.033} & \textbf{0.103} & 0.181 & \textbf{0.036} & \textbf{0.073} & 0.260 & \textbf{0.042} & \underline{0.088} \\
\textbf{VINS-FUSION}~\cite{Qin18} & stereo VO & 0.731 & 0.511 & 0.598 & 0.468 & 0.850 & 0.611 & 0.555 & 0.733 & 2.502 & 0.256 & 0.224 \\
\textbf{OKVIS2}~\cite{Leutenegger22} & stereo VO & \underline{0.115} & \textbf{0.070} & 0.115 & \underline{0.111} & 0.251 & \textbf{0.124} & 0.071 & \underline{0.084} & \underline{0.206} & \underline{0.047} & \textbf{0.071} \\
\hline
\textbf{BASALT}~\cite{Usenko19} & stereo VIO  & \underline{0.068} & \underline{0.065} & 0.062 & \underline{0.062} & 0.114 & 0.145 & \textbf{0.043} & \textbf{0.045} & \underline{0.054} & \underline{0.039} & \underline{0.049} \\
\textbf{*ORB-SLAM3}~\cite{Campos21}  & stereo VIO  & 0.121 & \textbf{0.037} & \textbf{0.043} & \textbf{0.038} & \underline{0.104} & \underline{0.110} & 0.087 & 0.144 & 0.474 & 0.066 & 0.103\\
\textbf{VINS-FUSION}~\cite{Qin18} & stereo VIO & 0.213 & 0.267 & 0.223 & 0.306 & 0.441 & 0.314 & 0.112 & 0.101 & 0.114 & 0.126 & 0.121 \\
\textbf{OKVIS2}~\cite{Leutenegger22} & stereo VIO & \textbf{0.063} & 0.072 & \underline{0.054} & 0.108 & \textbf{0.102} & \textbf{0.090} & 0.054 & 0.053 & \textbf{0.030} & 0.043 & \textbf{0.026} \\
\textbf{OpenVINS}~\cite{Geneva20} & stereo VIO & 0.144 & 0.084 & 0.095 & 0.165 & 0.295 & 0.517 & 0.061 & 0.055 & 0.056 & 0.062 & 0.052 \\
\textbf{SchurVINS}~\cite{Fan_2024_CVPR} & stereo VIO & 0.086 & 0.076 & 0.067 & 0.097 & 0.183 & 0.138 & \underline{0.048} & \underline{0.051} & 0.083 & \textbf{0.036} & 0.078 \\
\textbf{*ROVIO}~\cite{Bloesch17} & mono VIO & 0.433 & 0.267 & \textit{fail} & 0.383 & 0.785 & 1.289 & 0.165 & 0.182 & 0.192 & 0.243 & 0.393 \\
\hline
\hline
\textbf{milliseconds/frame} & \textbf{Method}  & \textbf{Avg} & \textbf{MH01} & \textbf{MH02} & \textbf{MH03} & \textbf{MH04} & \textbf{MH05} & \textbf{V101} & \textbf{V102} & \textbf{V103} & \textbf{V201} & \textbf{V202}   \\
\hline
\textbf{SMF-VO (ours)} & stereo VO & \textbf{7.789} & \textbf{5.419} & \textbf{5.949} & \textbf{7.344} & \textbf{8.317} & \textbf{8.041} & \textbf{6.088} & \textbf{9.355} & \textbf{10.839} & \textbf{6.096} & \textbf{10.443} \\
\textbf{BASALT}~\cite{Usenko19} & stereo VO & \underline{23.706} & \underline{29.101} & \underline{27.447} & \underline{28.231} & \underline{26.690} & \underline{25.990} & \underline{23.980} & \underline{20.473} & \underline{16.482} & \underline{20.597} & \underline{18.065} \\
\textbf{*ORB-SLAM3}~\cite{Campos21}  & stereo VO & 65.674 & 83.890 & 72.340 & 72.568 & 64.944 & 67.565 & 61.035 & 55.170 & 52.934 & 54.699 & 71.585 \\
\textbf{VINS-FUSION}~\cite{Qin18} & stereo VO  & 51.506 & 51.304 & 51.023 & 50.799 & 51.070 & 51.062 & 51.007 & 55.836 & 51.817 & 50.560 & 50.583 \\
\textbf{OKVIS2}~\cite{Leutenegger22} & stereo VO & 127.902 & 167.920 & 161.468 & 157.856 & 129.044 & 132.349 & 121.978 & 105.822 & 86.944 & 111.945 & 103.690 \\
\hline
\textbf{BASALT}~\cite{Usenko19} & stereo VIO & 20.032 & 23.199 & \underline{23.146} & 22.403 & 21.743 & 22.169 & 20.961 & 17.201 & \underline{13.770} & 19.315 & \underline{16.414} \\
\textbf{\textbf{*}ORB-SLAM3}~\cite{Campos21}  & stereo VIO & 65.467 & 76.937 & 77.294 & 82.313 & 64.680 & 67.881 & 57.988 & 53.220 & 52.451 & 53.848 & 68.059 \\
\textbf{VINS-FUSION}~\cite{Qin18} & stereo VIO & 51.506 & 51.366 & 51.055 & 50.815 & 51.100 & 51.046 & 51.003 & 55.785 & 51.741 & 50.591 & 50.561 \\
\textbf{OKVIS2}~\cite{Leutenegger22} & stereo VIO & 128.576 & 173.060 & 163.651 & 157.079 & 124.421 & 127.595 & 124.556 & 108.578 & 86.777 & 113.889 & 106.154 \\
\textbf{OpenVINS}~\cite{Geneva20} & stereo VIO  & 50.159 & 32.704 & 40.020 & 45.321 & 65.177 & 54.182 & 39.219 & 68.862 & 56.722 & 50.200 & 49.185 \\
\textbf{SchurVINS}~\cite{Fan_2024_CVPR} & stereo VIO & \textbf{10.347} & \textbf{10.815} & \textbf{10.659} & \textbf{10.701} & \textbf{10.623} & \textbf{10.651} & \textbf{10.019} & \textbf{9.997} & \textbf{9.858} & \textbf{9.976} & \textbf{10.174} \\
\textbf{*ROVIO}~\cite{Bloesch17} & mono VIO & \underline{18.893} & \underline{15.775} & \textit{fail} & \underline{17.702} & \underline{16.049} & \underline{18.811} & \underline{16.566} & \underline{14.533} & 16.190 & \underline{15.424} & 16.971 \\
\hline
\end{tabular}%
}
\caption{\textbf{RMSE ATE (m) and execution time (ms/frame) on the EuRoC dataset} using a Raspberry Pi 5 (2.4~GHz quad-core). The V203 sequence contains many featureless frames, causing failures in most odometry systems, and is therefore excluded. Refer to the text for the details on ROVIO and ORB-SLAM3. 
The best result is shown in \textbf{bold}, and the second-best is \underline{underlined}. Entries marked ``\textit{fail}'' indicate RMSE ATE values exceeding 100~m. Our system achieves the fastest average speed and can operate at 100~Hz on lightweight embedded platforms.
}
\label{tab:euroc_time_ate_all}
\end{table*}

\begin{table*}[tb!]
\centering
\resizebox{\linewidth}{!}{
\begin{tabular}{| r c | r | r r r r r r r r r r |}
\hline
\textbf{RMSE ATE (m)} & \textbf{Method} & \textbf{Avg} & \textbf{00} & \textbf{02} & \textbf{03} & \textbf{04} & \textbf{05} & \textbf{06} & \textbf{07} & \textbf{08} & \textbf{09} & \textbf{10}  \\
\hline
\textbf{SMF-VO (ours)} & stereo VO & \underline{2.886} & \textbf{3.833} & \textbf{5.604} & 1.519 & 1.561 & \underline{2.490} & 3.568 & \textbf{1.346} & \textbf{3.557} & 4.313 & \underline{1.074} \\
\textbf{BASALT}~\cite{Usenko19} & stereo VO & 3.267 & \underline{3.864} & 10.594 & \underline{1.330} & \underline{1.289} & 2.930 & \underline{2.532} & 1.403 & 3.766 & \underline{3.845} & 1.114 \\
\textbf{*ORB-SLAM3}~\cite{Campos21}  & stereo VO & \textbf{2.668} & 4.811 & \underline{8.122} & \textbf{0.776} & \textbf{0.218} & \textbf{1.840} & \textbf{1.834} & \underline{1.383} & \underline{3.558} & \textbf{3.078} & \textbf{1.062} \\
\textbf{VINS-FUSION}~\cite{Qin18} & stereo VO & 7.241 & 12.295 & 22.353 & 1.862 & 1.553 & 6.189 & 3.838 & 2.210 & 10.388 & 8.125 & 3.601 \\
\hline
\hline
\textbf{milliseconds/frame} & \textbf{Method} & \textbf{Avg} & \textbf{00} & \textbf{02} & \textbf{03} & \textbf{04} & \textbf{05} & \textbf{06} & \textbf{07} & \textbf{08} & \textbf{09} & \textbf{10} \\
\hline
\textbf{SMF-VO (ours)} & stereo VO & \textbf{19.348} & \textbf{19.411} & \textbf{21.469} & \textbf{13.874} & \textbf{21.169} & \textbf{19.691} & \textbf{21.214} & \textbf{16.818} & \textbf{19.382} & \textbf{21.937} & \textbf{18.513} \\
\textbf{BASALT}~\cite{Usenko19} & stereo VO & \underline{32.047} & \underline{32.060} & \underline{35.625} & \underline{33.121} & \underline{30.295} & \underline{30.851} & \underline{30.745} & \underline{31.689} & \underline{31.604} & \underline{31.666} & \underline{32.814} \\
\textbf{*ORB-SLAM3}~\cite{Campos21}  & stereo VO & 88.311 & 85.430 & 86.354 & 91.123 & 97.618 & 86.092 & 96.072 & 86.693 & 85.691 & 86.418 & 81.616 \\
\textbf{VINS-FUSION}~\cite{Qin18} & stereo VO & 98.301 & 96.560 & 97.743 & 105.830 & 101.845 & 97.762 & 94.323 & 99.328 & 97.359 & 97.436 & 94.821\\
\hline
\end{tabular}%
}
\caption{\textbf{RMSE ATE (m) and execution time (ms/frame) on the KITTI dataset} using a Raspberry Pi 5 (2.4~GHz quad-core). The 01 sequence contains many featureless frames and dynamic objects, causing failures in most odometry systems, and is therefore excluded. Refer to the text for the details on ORB-SLAM3. 
The best result is shown in \textbf{bold}, and the second-best is \underline{underlined}.
}
\label{tab:kitti_time_ate_all}
\end{table*}

\begin{table*}[tb!]
\centering
\resizebox{\linewidth}{!}{
\begin{tabular}{| r c || r | r r r r r r || r | r r r r r r |}
\hline
 & & \multicolumn{7}{c||}{\textbf{RMSE ATE (m)}} & \multicolumn{7}{c|}{\textbf{milliseconds/frame}}  \\
\cline{3-16}
 & \textbf{Method} & \textbf{Avg} & \textbf{R1} & \textbf{R2} & \textbf{R3} & \textbf{R4} & \textbf{R5} & \textbf{R6}  & \textbf{Avg} & \textbf{R1} & \textbf{R2} & \textbf{R3} & \textbf{R4} & \textbf{R5} & \textbf{R6} \\
\hline
\textbf{SMF-VO (ours)} & stereo VO & \underline{0.082} & \textbf{0.053} & 0.183 & \underline{0.114} & \underline{0.036} & \textbf{0.059} & \underline{0.048} & \textbf{9.733} & \textbf{11.501} & \textbf{8.551} & \textbf{8.231} & \textbf{10.077} & \textbf{11.201} & \textbf{8.836} \\
\textbf{BASALT}~\cite{Usenko19} & stereo VO & 0.205 & 0.112 & \underline{0.128} & 0.157 & 0.039 & 0.758 & \textbf{0.034} & \underline{14.342} & \underline{14.374} & \underline{13.756} & \underline{13.542} & \underline{14.690} & \underline{13.489} & \underline{16.200} \\
\textbf{*ORB-SLAM3}~\cite{Campos21}  & stereo VO & \textbf{0.079} & \underline{0.078} & \textbf{0.052} & \textbf{0.099} & 0.082 & \underline{0.077} & 0.086 & 79.965 & 77.761 & 77.387 & 78.585 & 85.519 & 82.275 & 78.264 \\
\textbf{OKVIS2}~\cite{Leutenegger22} & stereo VO & 0.104 & 0.085 & 0.134 & 0.169 & \textbf{0.021} & 0.142 & 0.070 & 94.768 & 97.108 & 94.894 & 93.885 & 91.408 & 96.467 & 94.845 \\
\hline
\textbf{BASALT}~\cite{Usenko19} & stereo VIO & 0.095 & 0.108 & \textbf{0.076} & 0.133 & 0.052 & 0.180 & \textbf{0.020} & \textbf{19.344} & \textbf{19.235} & \textbf{18.581} & \textbf{17.404} & \textbf{20.501} & \textbf{17.317} & \textbf{23.024} \\
\textbf{*ORB-SLAM3}~\cite{Campos21}  & stereo VIO & \textbf{0.044} & \textbf{0.038} & \underline{0.102} & \textbf{0.059} & \textbf{0.010} & \textbf{0.011} & \underline{0.043} & 72.636 & 70.178 & 73.244 & 71.614 & 72.932 & 74.797 & 73.051  \\
\textbf{OKVIS2}~\cite{Leutenegger22} & stereo VIO  & \underline{0.067} & \underline{0.054} & 0.124 & \underline{0.099} & 0.042 & \underline{0.038} & 0.048 & 91.096 & 95.178 & 91.234 & 90.934 & 86.515 & 90.585 & 92.129 \\
\textbf{OpenVINS}~\cite{Geneva20} & stereo VIO & 0.081 & 0.072 & 0.139 & 0.106 & \underline{0.030} & 0.079 & 0.057 & \underline{43.083} & \underline{41.156} & \underline{40.382} & \underline{41.687} & \underline{51.948} & \underline{41.465} & \underline{41.862} \\
\hline

\end{tabular}%
}
\caption{\textbf{RMSE ATE(m) and Execution time (ms/frame) on the TUM-VI Room dataset} using a Raspberry Pi 5 (2.4 GHz quad-core). 
 Refer to the text for the details on ORB-SLAM3. 
 The best value is shown in \textbf{bold}, and the second best is \underline{underlined}.}
\label{tab:tumvi_time_ate_all}
\end{table*}

\begin{figure*}[tb!]
    \centering
    \begin{minipage}{0.24\textwidth}
        \centering
        \includegraphics[width=\linewidth]{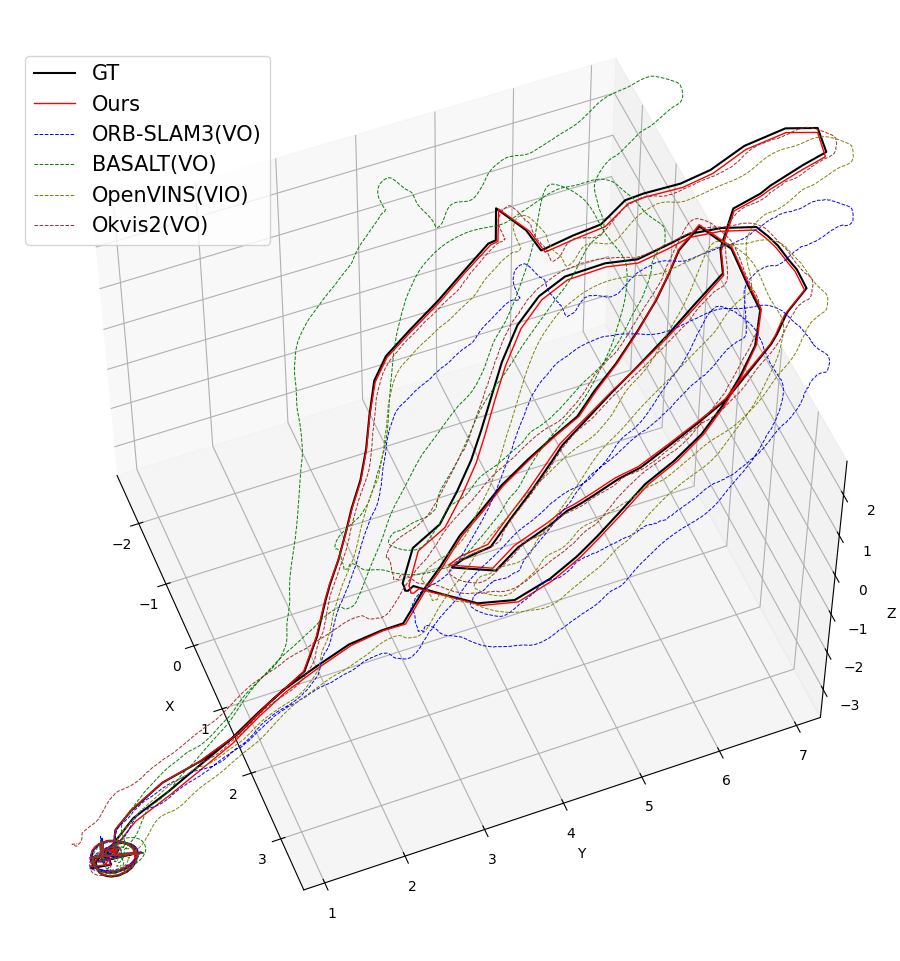}
        \vspace{-0.5cm}
        \\ {\scriptsize (a) MH02 easy}
    \end{minipage}
    \hfill
    \begin{minipage}{0.24\textwidth}
        \centering
        \includegraphics[width=\linewidth]{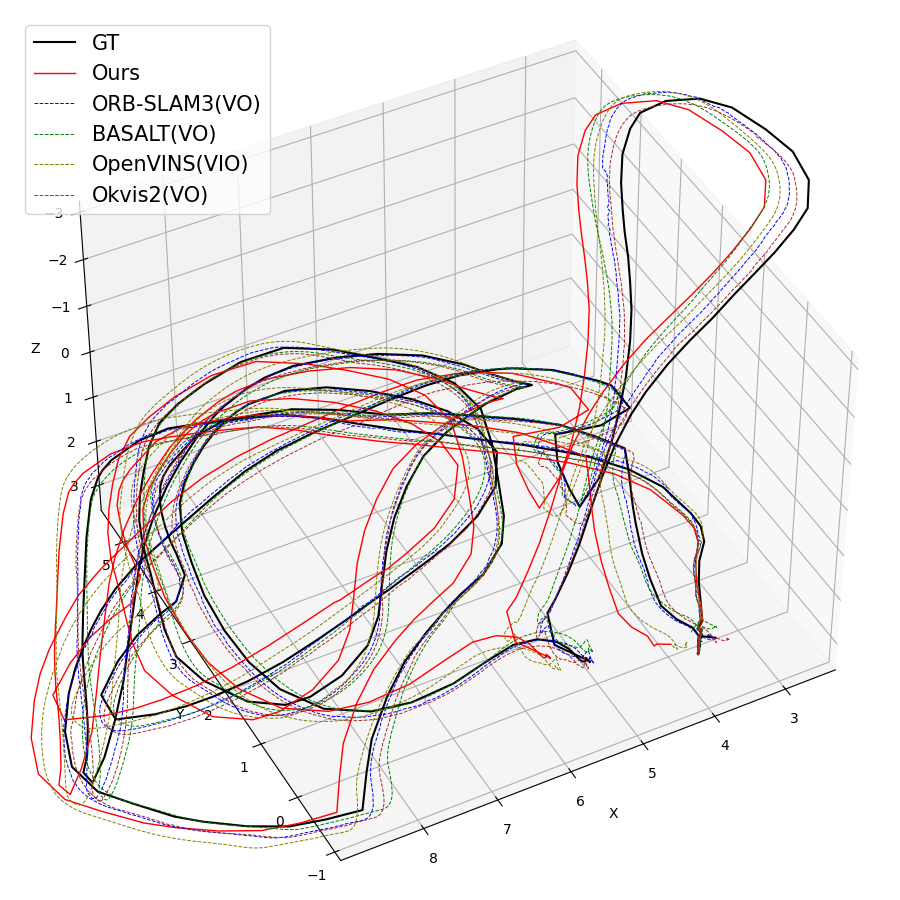}
        \vspace{-0.5cm}
        \\ {\scriptsize (b) MH03 medium}
    \end{minipage}
    \hfill
    \begin{minipage}{0.24\textwidth}
        \centering
        \includegraphics[width=\linewidth]{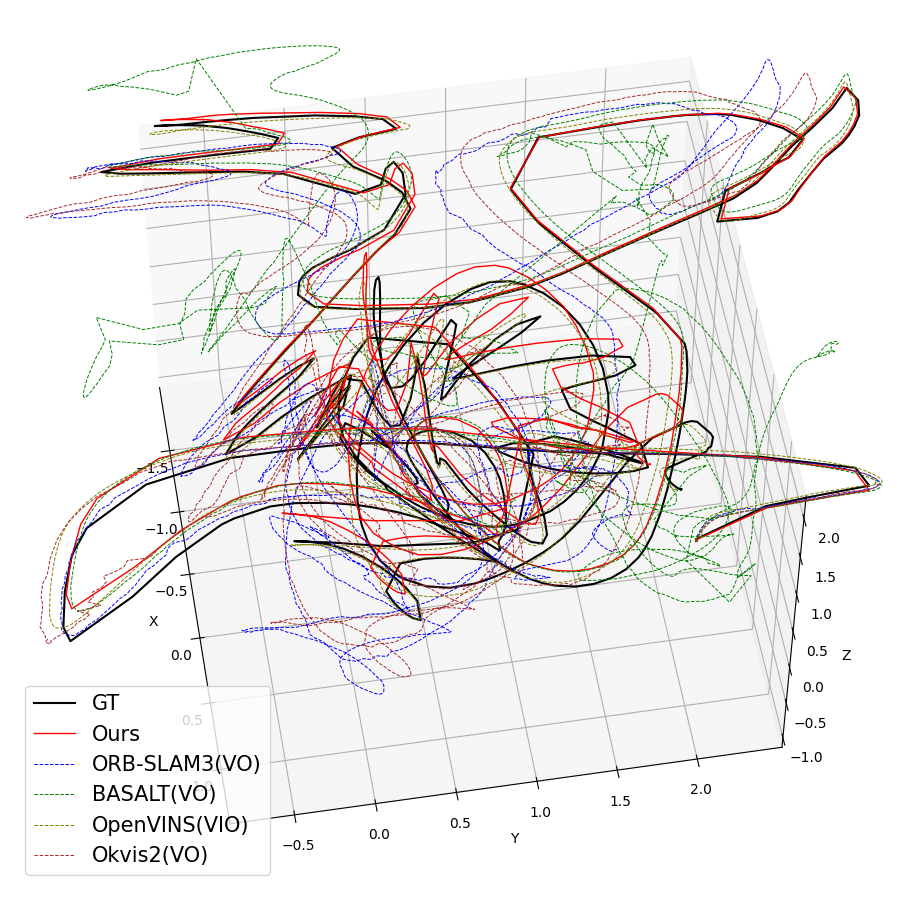}
        \vspace{-0.5cm}
        \\ {\scriptsize (c) V103 difficult}
    \end{minipage}
    \hfill
    \begin{minipage}{0.24\textwidth}
        \centering
        \includegraphics[width=\linewidth]{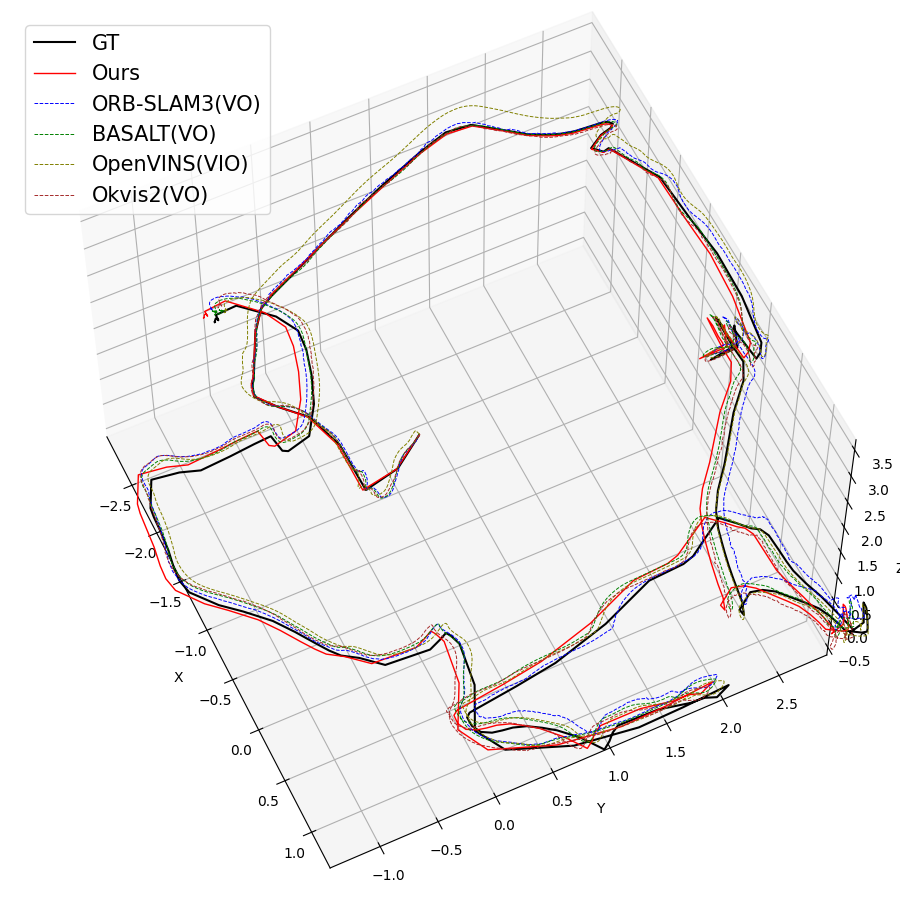}
        \vspace{-0.5cm}
        \\ {\scriptsize (c) V201 easy}
    \end{minipage}
    \vspace{-0.2cm}
    \caption{Qualitative trajectory result of EuRoC sequences. The estimated trajectories of SMF-VO (ours), ORB-SLAM3~\cite{Campos21} (VO), Basalt~\cite{Usenko19} (VO), OpenVINS~\cite{Geneva20} (VIO), and OKVIS2~\cite{Leutenegger22} (VO) are aligned to the ground truth pose at the first frame.}
    \label{fig:qual_euroc}
\end{figure*}

\begin{figure*}[tb!]
    \centering
    \begin{minipage}{0.24\textwidth}
        \centering
        \includegraphics[width=\linewidth]{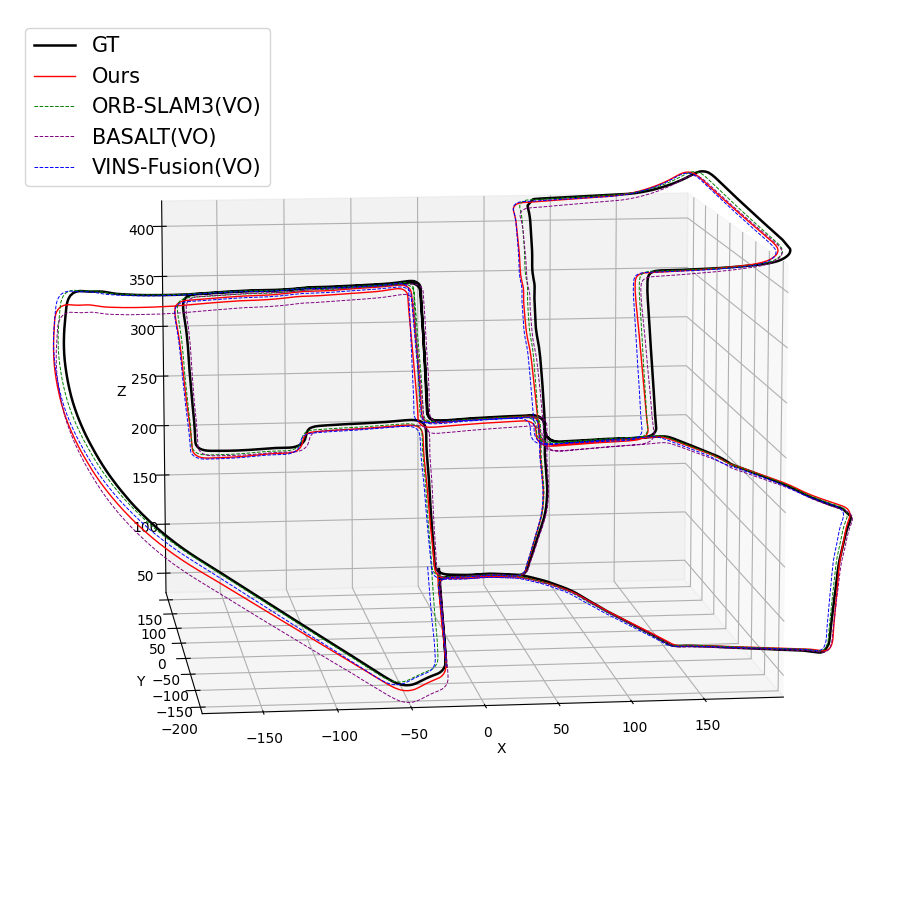}
        \vspace{-0.5cm}
        \\ {\scriptsize (a) kitti 00}
    \end{minipage}
    \hfill
    \begin{minipage}{0.24\textwidth}
        \centering
        \includegraphics[width=\linewidth]{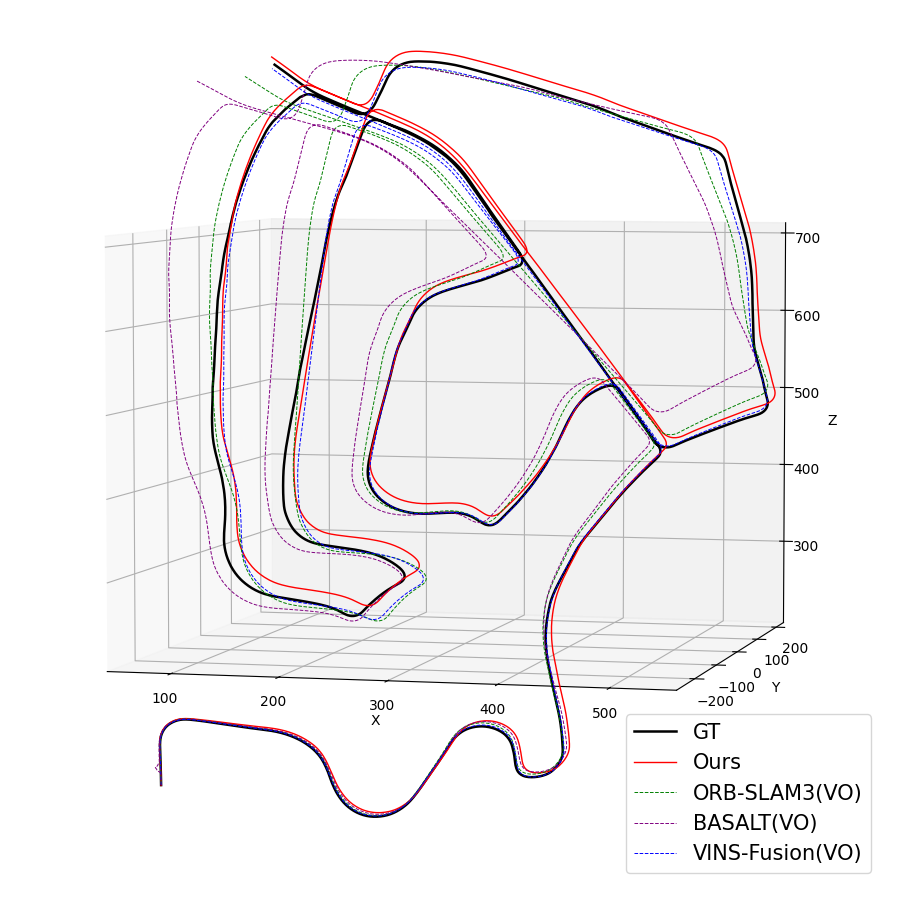}
        \vspace{-0.5cm}
        \\ {\scriptsize (b) kitti 02}
    \end{minipage}
    \hfill
    \begin{minipage}{0.24\textwidth}
        \centering
        \includegraphics[width=\linewidth]{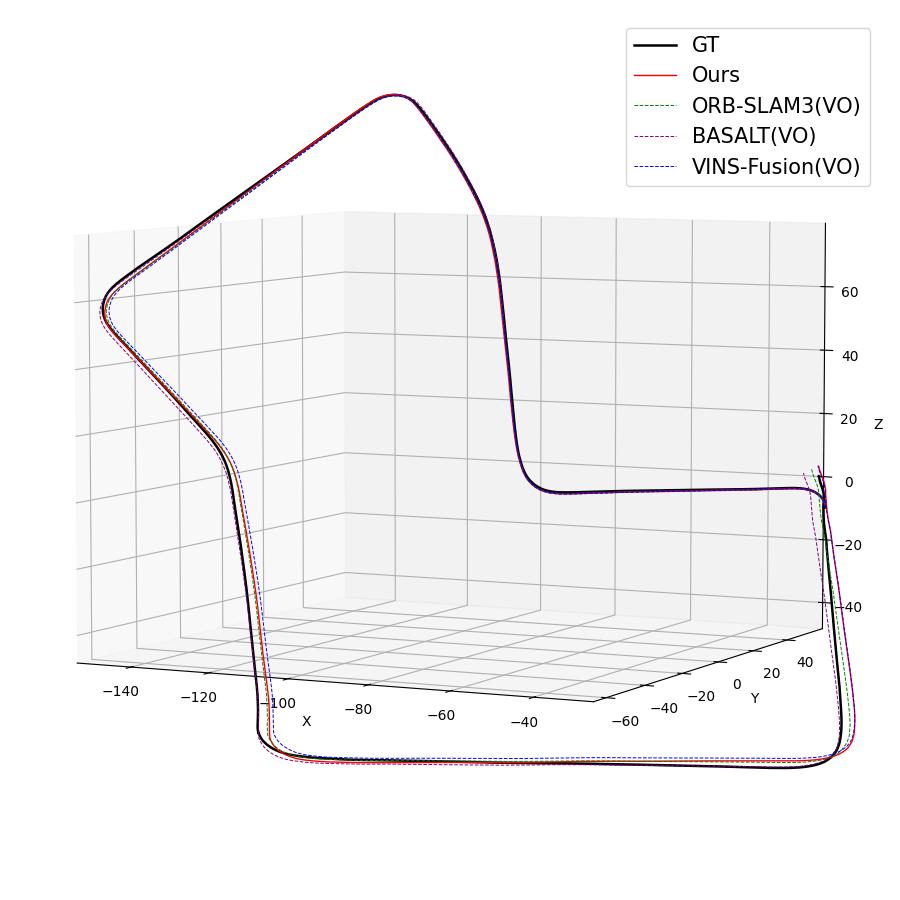}
        \vspace{-0.5cm}
        \\ {\scriptsize (c) kitti 07}
    \end{minipage}
    \hfill
    \begin{minipage}{0.24\textwidth}
        \centering
        \includegraphics[width=\linewidth]{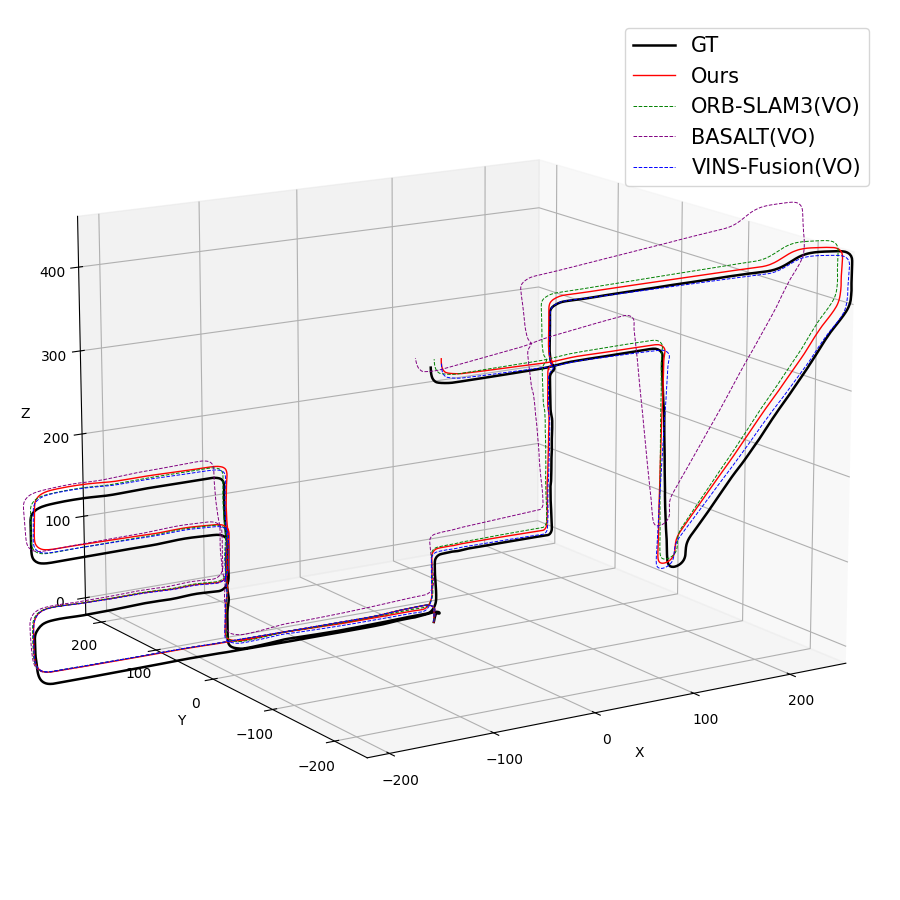}
        \vspace{-0.5cm}
        \\ {\scriptsize (c) kitti 08}
    \end{minipage}
    \vspace{-0.2cm}
    \caption{Qualitative trajectory result of KITTI sequences. The estimated trajectories of SMF-VO (ours), ORB-SLAM3~\cite{Campos21} (VO), Basalt~\cite{Usenko19} (VO), and VINS-Fusion~\cite{Qin18} (VO) are aligned to the ground truth pose at the first frame.}
    \label{fig:qual_kitti}
\end{figure*}

\begin{figure}[b!]
    \centering
    \begin{minipage}{0.236\textwidth}
        \centering
        \includegraphics[width=\linewidth]{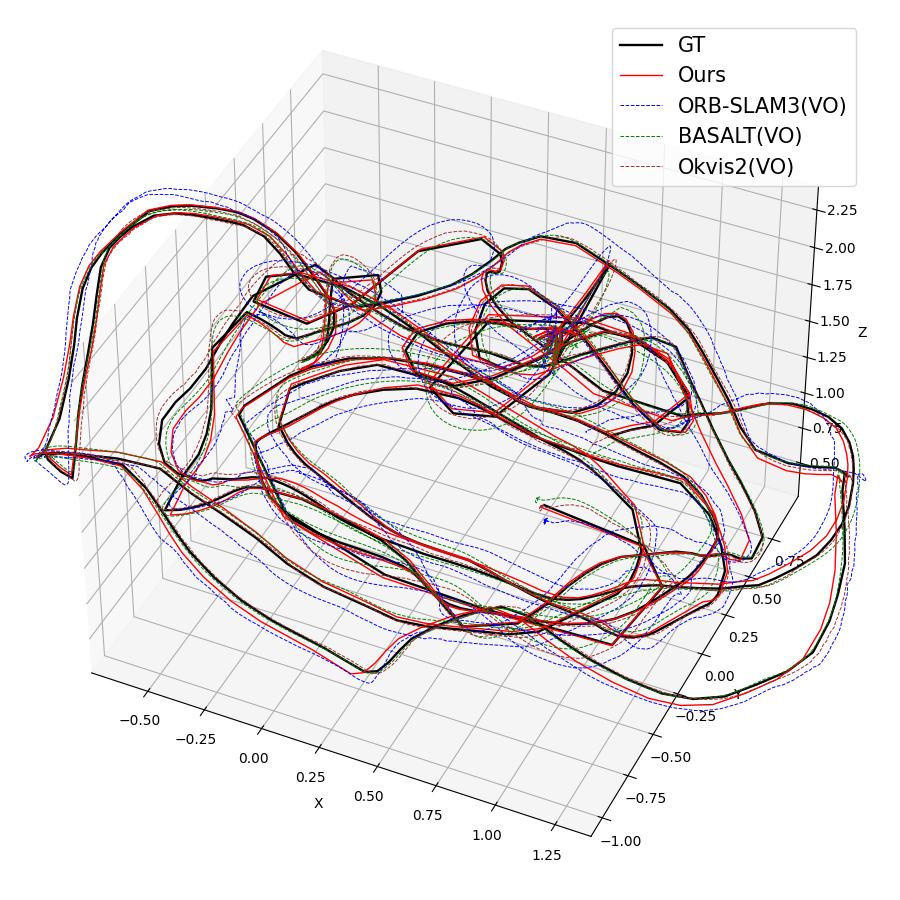} \\ 
        \vspace{-0.5cm}
        {\scriptsize (a) TUM-VI Room 4}
    \end{minipage}
    \hfill
    \begin{minipage}{0.236\textwidth}
        \centering
        \includegraphics[width=\linewidth]{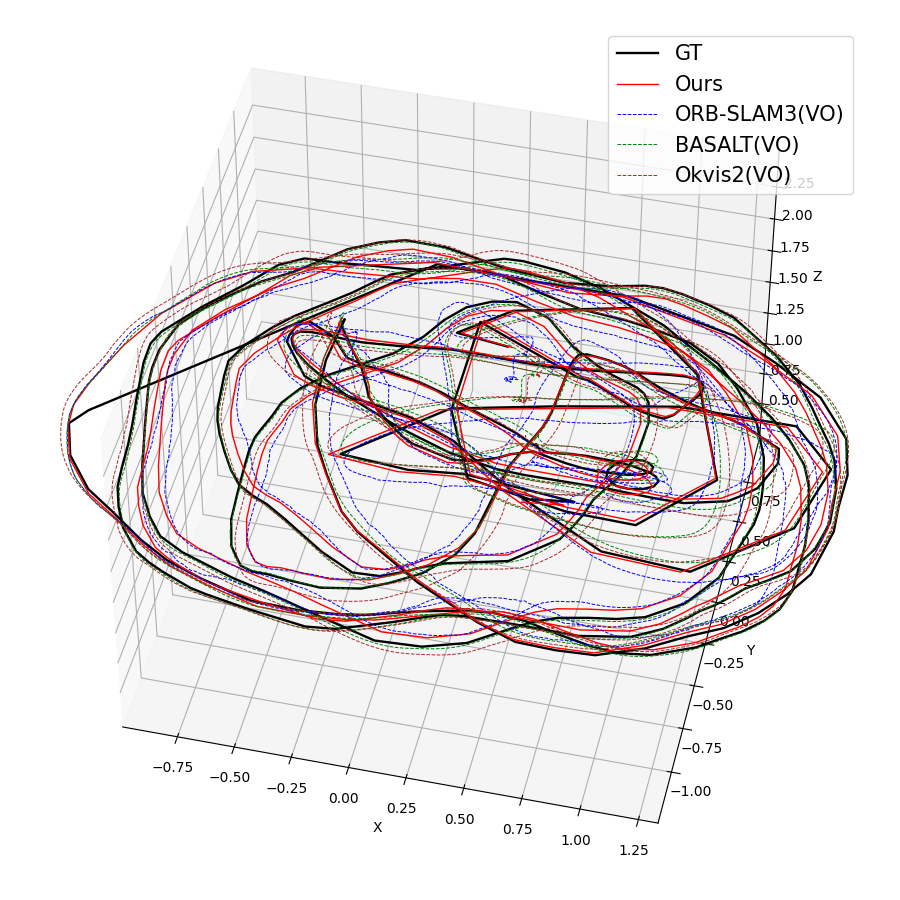}\\ 
        \vspace{-0.5cm}
        {\scriptsize (b) TUM-VI Room 6}
    \end{minipage}
    \vspace{-0.2cm}
    \caption{Qualitative trajectory results for TUM-VI Room sequences. The estimated trajectories of SMF-VO (ours), ORB-SLAM3~\cite{Campos21} (VO), Basalt~\cite{Usenko19} (VO), and OKVIS2~\cite{Leutenegger22} (VO) are aligned to the ground truth pose at the first frame.}
    \label{fig:qual_tumvi}
\end{figure}

We evaluate the proposed visual odometry method on three benchmark datasets: EuRoC~\cite{Burri16}, KITTI~\cite{Geiger12}, and TUM-VI~\cite{Schubert18}.
The EuRoC dataset, collected with a micro aerial vehicle, provides 11 stereo–IMU sequences.
KITTI, a standard benchmark for autonomous driving, offers 11 car-mounted sequences with GPS/INS ground truth.
TUM-VI contains fisheye stereo and IMU data under challenging motion; we use the Room sequences, as they are the only ones with full ground truth trajectories.

We compare our method against state-of-the-art VO/VIO algorithms under identical conditions.
All experiments run on a Raspberry Pi 5 (2.4 GHz quad-core ARM CPU) without GPU or hardware acceleration.
Each baseline is tested both in our setting (stereo without IMU) and in its recommended configuration (e.g., monocular or with IMU).
For ROVIO, only filter update time is measured. For ORB-SLAM3, we report tracking time excluding image preprocessing, which may make runtimes appear slightly faster.

Performance metrics are:
\begin{itemize}
\item Accuracy: RMSE of Absolute Trajectory Error (ATE).
\item Efficiency: Execution time on Raspberry Pi.
\end{itemize}
Finally, we conduct an ablation study to analyze the contribution of each component in the proposed SMF-VO framework.

\subsection{Evaluation on Benchmark Datasets}
Tables~\ref{tab:euroc_time_ate_all}, \ref{tab:kitti_time_ate_all}, and \ref{tab:tumvi_time_ate_all} report trajectory accuracy (RMSE ATE) and per-frame processing time for the EuRoC, KITTI, and TUM-VI Room datasets, emphasizing the efficiency of our method. For VO-only evaluation, loop closure is disabled in systems that provide it (e.g., ORB-SLAM3, VINS-FUSION, SchurVINS etc.). Qualitative trajectories are shown in Fig.~\ref{fig:qual_euroc},~\ref{fig:qual_kitti}, and~\ref{fig:qual_tumvi}, illustrating the robustness of our method against baselines and ground truth.

\paragraph{EuRoC}
On EuRoC (Table~\ref{tab:euroc_time_ate_all}), ORB-SLAM3 achieves the best VO-only accuracy with centimeter-level errors, benefiting from global bundle adjustment.  
Our method attains accuracy comparable to ORB-SLAM3, OKVIS2, and BASALT, with BASALT slightly ahead on some sequences due to local refinement.  
On average, SMF-VO ranks third, with an average error of approximately 0.1m.  

In efficiency, ORB-SLAM3 and OKVIS2 require 65$\sim$128 ms/frame on average, while SMF-VO runs in under 10 ms/frame, supporting operation at over 100 Hz on embedded boards.  
With IMU integration (VIO), only ORB-SLAM3, OKVIS2, and BASALT surpass SMF-VO in accuracy. BASALT particularly benefits from IMU data on challenging sequences such as V103.  


\paragraph{KITTI}
On KITTI (Table~\ref{tab:kitti_time_ate_all}), which provides only stereo images, ORB-SLAM3 again achieves the highest accuracy.  
Our method delivers similar accuracy with the lowest execution time. Compared to BASALT, the second-fastest VO system, SMF-VO is both more accurate and faster.  
Despite KITTI’s higher resolution and lower frame rate, our method processes frames in under 20 ms, ensuring real-time operation while remaining fast and efficient enough for deployment on embedded devices. These results confirm that SMF-VO generalizes well from small-scale indoor to large-scale outdoor environments, maintaining a favorable balance between accuracy and efficiency.

\paragraph{TUM-VI Room}
On TUM-VI Room (Table~\ref{tab:tumvi_time_ate_all}), which includes fisheye stereo, IMU, and ground-truth trajectories from motion capture, ORB-SLAM3 achieves the highest VO-only accuracy with centimeter-level errors.  
SMF-VO attains comparable accuracy, slightly surpassing ORB-SLAM3 on some sequences, and ranks second overall with an average error of about 0.08 m.  
In terms of efficiency, SMF-VO operates roughly 8$\times$ faster than ORB-SLAM3 and 1.5$\times$ faster than BASALT.  
Compared to VIO methods, SMF-VO remains highly competitive: it delivers higher accuracy than BASALT while requiring substantially less computation than the most accurate IMU-based systems, making it particularly attractive for embedded platforms.

\subsection{Ablation}

\begin{table}[!tb]
\footnotesize 
\centering
\resizebox{\linewidth}{!}{
\begin{tabular}{|r| r  r | r  r || r  r | r  r |}
\hline
 & \multicolumn{4}{c||}{\textbf{w/o nonlinear optimization}~\ref{sec:n-opt}} & \multicolumn{4}{c|}{\textbf{w/ nonlinear optimization}~\ref{sec:n-opt}} \\
\cline{2-9}
 & \multicolumn{2}{c|}{\textbf{pixel-based}~\ref{subsec:2DMF}} & \multicolumn{2}{c||}{\textbf{ray-based}~\ref{subsec:3dray}} & \multicolumn{2}{c|}{\textbf{pixel-based}~\ref{subsec:2DMF}} & \multicolumn{2}{c|}{\textbf{ray-based}~\ref{subsec:3dray}} \\
\cline{2-9}
 & ms/f & RMSE & ms/f & RMSE & ms/f & RMSE & ms/f & RMSE\\
\hline
\hline
\textbf{MH01} & 4.838 & 0.318 & 5.049 & \textbf{0.240} & 5.281 & \textbf{0.075} & 5.419 & 0.099\\
\textbf{MH02} & 5.030 & 0.287 & 4.957 & \textbf{0.214} & 5.955 & 0.099 & 5.949 & \textbf{0.039}\\
\textbf{MH03} & 4.914 & 1.070 & 4.987 & \textbf{0.783} & 7.285 & 0.173 & 7.344 & \textbf{0.169}\\
\textbf{MH04} & 5.092 & \textbf{5.491} & 5.052 & 5.722 & 8.948 & 0.297 & 8.317 & \textbf{0.208}\\
\textbf{MH05} & 4.784 & \textbf{3.277} & 5.090 & 3.386 & 7.889 & \textbf{0.267} & 8.041 & 0.348\\
\textbf{V101} & 4.914 & 0.235 & 4.840 & \textbf{0.187} & 5.986 & 0.083 & 6.088 & \textbf{0.063}\\
\textbf{V102} & 4.993 & \textbf{0.382} & 4.900 & 0.396 & 9.394 & 0.123 & 9.355 & \textbf{0.100}\\
\textbf{V103} & 5.026 & 0.644 & 4.916 & \textbf{0.628} & 11.322 & 0.097 & 10.839 & \textbf{0.088}\\
\textbf{V201} & 4.818 & \textbf{0.184} & 4.854 & 0.281 & 6.045 & 0.064 & 6.096 & \textbf{0.057}\\
\textbf{V202} & 5.166 & 0.511 & 4.966 & \textbf{0.186} & 10.344 & 0.125 & 10.443 & \textbf{0.114}\\
\hline
\textbf{Avg} & 4.957 & 1.240 & 4.961 & \textbf{1.202} & 7.845 & 0.140 & 7.789 & \textbf{0.128}\\
\hline
\hline
\textbf{Room1} & 4.236 & 4.339 & 3.805 & \textbf{0.897} & 10.897 & 0.111 & 11.501 & \textbf{0.053}\\
\textbf{Room2} & 3.866 & 1.916 & 3.840 & \textbf{1.212} & 9.147 & 0.209 & 8.551 & \textbf{0.183}\\
\textbf{Room3} & 4.414 & 1.505 & 3.819 & \textbf{1.089} & 9.719 & 0.136 & 8.231 & \textbf{0.114}\\
\textbf{Room4} & 3.919 & 1.261 & 3.780 & \textbf{0.251} & 9.741 & 0.041 & 10.077 & \textbf{0.036}\\
\textbf{Room5} & 4.475 & 4.823 & 3.997 & \textbf{0.754} & 10.770 & 0.103 & 11.201 & \textbf{0.059}\\
\textbf{Room6} & 3.658 & 0.631 & 3.753 & \textbf{0.320} & 8.725 & 0.051 & 8.836 & \textbf{0.048}\\
\hline
\textbf{Avg} & 4.095 & 2.413 & 3.832 & \textbf{0.754} & 9.833 & 0.108 & 9.733 & \textbf{0.082}\\
\hline
\end{tabular}
}
\caption{
Ablation studies were performed on pixel-based and ray-based motion field algorithms, both with and without the proposed lightweight nonlinear optimization step.
The values with better accuracy (RMSE ATE (m)) in each setup (w/ and w/o nonlinear optimization) are shown in \textbf{bold}.
The results highlight the benefits of the proposed ray-based approach and the effectiveness of the lightweight nonlinear optimization.
}
\label{tab:ablation_new}
\end{table}

{Ablation studies were conducted to compare the performance of the ray-based motion field (Section~\ref{subsec:3dray}) with the conventional pixel-based motion field (Section~\ref{subsec:2DMF}), and to determine the effect of adding a lightweight nonlinear optimization step after RANSAC-based motion estimation.}  

As shown in Table~\ref{tab:ablation_new}, the 3D ray-based formulation consistently improves trajectory accuracy, particularly on wide field-of-view data such as the TUM-VI Room sequences. The performance gap becomes more pronounced when nonlinear optimization is omitted. This is because, in the 2D pixel-based formulation, the error bound on pixel locations grows with distance from the image center, making the linear system $\bA\bx = \bb$ less reliable and more sensitive to outliers. In contrast, the 3D ray-based formulation maintains a consistent angular error metric—measuring deviations in degrees between observed and reprojected rays—regardless of image distortion or projection model. This property makes it more robust and better suited to wide FoV or distorted cameras, whereas the 2D pixel-based method assumes an ideal pinhole model.   

Regarding runtime, omitting the nonlinear optimization step reduces computation further, while maintaining acceptable accuracy on sequences with simple motion and sufficient texture (e.g., MH01, MH02, V101, V201). Although the optimization step significantly improves accuracy in challenging cases, in structured and less dynamic environments it can be skipped to achieve faster execution with only a modest loss in accuracy.


\section{Conclusion}

In this paper, we introduce SMF-VO, a novel motion-centric visual odometry framework that directly estimates ego-motion from sparse optical flow using the motion field equation. Unlike conventional VO methods that rely on explicit pose estimation and computationally expensive map optimization, SMF-VO computes instantaneous linear and angular velocities from a sparse motion field using a highly efficient linear solution, complemented by an optional, lightweight nonlinear refinement step.

\IGNORE{A key contribution of our work is the generalized 3D ray-based motion field formulation, which overcomes the limitations of traditional 2D pixel-based methods that are confined to the pinhole projection model and cannot handle wide-field-of-view cameras like fisheye lenses.} 
Through extensive experiments on standard benchmark datasets, we demonstrated that SMF-VO significantly outperforms conventional VO approaches in computational efficiency while maintaining competitive accuracy. Our algorithm achieved over 100 FPS on a Raspberry Pi 5 using only a CPU, highlighting its suitability for real-time applications on low-power embedded platforms.

In conclusion, SMF-VO establishes a new, motion-centric state estimation paradigm that offers a scalable and efficient alternative to conventional VO. Its exceptional computational efficiency makes it a strong candidate for practical, real-world applications in mobile robotics, AR/VR, and wearable devices. Future work will explore its application to diverse sensor modalities, including event cameras and IMU fusion.





\addtolength{\textheight}{-4cm}   



\IGNORE{
\section*{APPENDIX}
Appendixes should appear before the acknowledgment.
}

\IGNORE{
\section*{ACKNOWLEDGMENT}
This work was supported  
in part by (---),
%
in part by (---).
}
\bibliographystyle{IEEEtran}
\bibliography{root}

\end{document}